\documentclass{ecai}
\usepackage{times}
\usepackage{graphicx}
\usepackage{latexsym}

\usepackage{booktabs}
\usepackage{amsmath}

\begin{document}

\title{Learning to Reuse Translations: Guiding Neural Machine Translation with Examples}

\author{Qian Cao\textsuperscript{1}, Shaohui Kuang\textsuperscript{1} \and Deyi Xiong\institute{Soochow University,
China, email: qcao@stu.suda.edu.cn, shaohuikuang@foxmail.com, dyxiong@suda.edu.cn}  }

\maketitle
\bibliographystyle{ecai}

\begin{abstract}
  In this paper, we study the problem of enabling neural machine translation (NMT) to reuse previous translations from similar examples in target prediction. Distinguishing reusable translations from noisy segments and learning to reuse them in NMT are non-trivial. To solve these challenges, we propose an Example-Guided NMT (EGNMT) framework with two models: (1) a noise-masked encoder model that masks out noisy words according to word alignments and encodes the noise-masked sentences with an additional example encoder and (2) an auxiliary decoder model that predicts reusable words via an auxiliary decoder sharing parameters with the primary decoder. We define and implement the two models with the state-of-the-art Transformer. Experiments show that the noise-masked encoder model allows NMT to learn useful information from examples with low fuzzy match scores (FMS) while the auxiliary decoder model is good for high-FMS examples. More experiments on Chinese-English, English-German and English-Spanish translation demonstrate that the combination of the two EGNMT models can achieve improvements of up to +9 BLEU points over the baseline system and +7 BLEU points over a two-encoder Transformer.
\end{abstract}

\section{Introduction}

Neural machine translation \cite{bahdanau2015neural,sutskever2014sequence,vaswani2017attention,gehring2017convolutional} captures the knowledge of the source and target language along with their correspondences as part of the encoder and decoder parameters learned from data. With this embedded and parameterized knowledge, a trained NMT model is able to translate a new source sentence into the target language.

In this paper, we consider a different translation scenario to NMT. In this scenario, in addition to a given source sentence, NMT is also provided with an example translation that contains reusable translation segments for the source sentence. The NMT model can either use the embedded knowledge in parameters or learn from the example translation on the fly to predict target words. This translation scenario is not new to machine translation as it has been studied in example-based machine translation \cite{nagao1984framework} and the combination of statistical machine translation (SMT) with translation memory \cite{koehn2010convergence}. However, in the context of NMT, the incorporation of external symbol translations is still an open problem. We therefore propose example-guided NMT (EGNMT) to seamlessly integrate example translations into NMT.

Unlike conventional machine translation formalisms, a trained NMT model is not easy to be quickly adapted to an example translation as the model is less transparent and amenable than SMT models. To address this issue, we use a new encoder (thereafter the example encoder) to encode the example translation in EGNMT, in addition to the primary encoder for the source sentence.

As the example is not identical to the source sentence, only parts of the example translation can be used in the final translation for the source sentence. Hence the challenge is to teach EGNMT to detect and use matched translation fragments while ignoring unmatched noisy parts.

To handle this challenge, we propose two models that guide the decoder to reuse translations from examples. The first model is a noise-masked encoder model (NME). 
In the example encoder, we pinpoint unmatched noisy fragments in each example translation via word alignments and mask them out with a symbol ``$\langle X \rangle$''. The noise-masked example translation is then input to the example encoder. This model mimics human translators in paying special attention to reusable parts and ignoring those unrelated parts when an example translation is given.

Different from NME that encodes the noise-masked example translation, in the second model, we directly produce a masked translation from the example translation with an auxiliary decoder (hence the auxiliary decoder model, AD). We compare the reference translation of a source sentence in the training data with its corresponding example translation. The identical parts in the reference translation are retained while other parts are substituted with the symbol ``$\langle X \rangle$''. The auxiliary decoder is then used to predict the masked reference translation. It is jointly trained with the primary decoder and shares its parameters with the primary decoder. Therefore the primary decoder can learn from the auxiliary decoder to predict reusable words/phrases from the example translation. Notice that the auxiliary decoder is only used during the joint training phase.

In summary, our contributions are threefold.
\begin{itemize}
\item We propose an example-guided NMT framework to learn to reuse translations from examples.
\item In this framework, we further propose two models: NME that encodes reusable translations in the example encoder and AD that teaches the primary decoder to directly predict reusable translations with the auxiliary decoder via parameter sharing and joint training. 
\item The proposed EGNMT framework can be used to any encoder-decoder based NMT. In this paper, we define EGNMT over the state-of-the-art NMT architecture Transformer \cite{vaswani2017attention} and evaluate EGNMT on Chinese-English, English-German and English-Spanish translation. In our experiments, the best EGNMT model achieves improvements of 4-9 BLEU points over the baseline on the three language pairs. Analyses show that the proposed model can effectively learn from example translations with different similarity scores.
\end{itemize}

\section{Related Work}

{\bf Translation Memory} Our work is related to the studies that combine translation memory (TM) with machine translation. Various approaches have been proposed for the combination of TM and SMT. For example, Koehn and Senellart \cite{koehn2010convergence} propose to reuse matched segments from TM for SMT. In NMT, Gu et al. \cite{gu2017search} propose to encode sentences from TM into vectors, which are then stored as key-value pairs to be explored by NMT. Cao and Xiong \cite{cao2018encoding} regard the incorporation of TM into NMT as a multi-input problem and use a gating mechanism to combine them. Bapna and Firat \cite{bapna2019non} integrate  multiple similar examples into NMT and explore different retrieval strategies. Different from these methods, we propose more fine-grained approaches to dealing with noise in matched translations.

{\bf Example-based MT} In the last century, many studies have focused on the impact of examples on translation, or translation by analogy \cite{nagao1984framework,somers1999example}. Wu \cite{wu2005mt} discuss the relations of statistical, example-based and compositional MT in a three-dimensional model space because of the interplay of them. Our work can be considered as a small step in this space to integrate the example-based translation philosophy with NMT.

{\bf Using examples in neural models for other tasks} In other areas of natural language processing, many researchers are interested in combining symbolic examples with neural models. Pandey et al. \cite{pandey2018exemplar} propose a conversational model that learns to utilize similar examples to generate responses. The retrieved examples are used to create exemplar vectors that are used by the decoder to generate responses. Cai et al. \cite{cai2018skeleton} also introduce examples into dialogue systems, but they first generate a skeleton based on the retrieved example, and then use the skeleton to serve as an additional knowledge source for response generation. Guu et al. \cite{guu2018generating} present a new generative language model for sentences that first samples a prototype sentence and then edits it into a new sentence.

{\bf External knowledge for NMT} Our work is also related to previous works that incorporate external knowledge or information into NMT. Zhou et al. \cite{zhou2017neural} propose to integrate the outputs of SMT to improve the translation quality of NMT while Wang et al. \cite{wang2017neural} explore SMT recommendations in NMT. Zhang et al. \cite{zhang2018guiding} incorporate translation pieces into NMT within beam search. In document translation, many efforts try to encode the global context information by the aid of discourse-level approaches \cite{kuang2018fusing,zhang2018improving,voita2018context}. In addition to these, some studies integrate external dictionaries into NMT \cite{arthur2016incorporating,li2016towards} or force the NMT decoder to use given words/phrases in target translations \cite{hokamp2017lexically,post2018fast,hasler2018neural}.

{\bf Multi-task learning} The way that we use the auxiliary decoder and share parameters is similar to multi-task learning in NMT. Just to name a few, Dong et al. \cite{dong2015multi} share an encoder among different translation tasks. Weng et al. \cite{weng2017neural} add a word prediction task in the process of translation. Sachan and Neubig \cite{sachan2018parameter} explore the parameter sharing strategies for the task of multilingual machine translation. Wang et al. \cite{wang2018learning} propose to jointly learn to translate and predict dropped pronouns.

\section{Guiding NMT with Examples}

The task here is to translate a source sentence into the target language from the representations of the sentence itself and a matched example translation. In this section, we first introduce how example translations are retrieved and then briefly describe the basic EGNMT model that uses one encoder for source sentences and the other for retrieved example translations. Based on this simple model, we elaborate the proposed two models: the noise-masked encoder model and auxiliary decoder model.

\subsection{Example Retrieval}

Given a source sentence $x$ to be translated, we find a matched example $(x^m, y^m)$ from an example database $D = \{(x_i,y_i)\}^N_1$ with $N$ source-target pairs. The source part $x^m$ of the matched example has the highest similarity score to $x$ in $D$.  A variety of metrics can be used to estimate this similarity score. In this paper, we first get the top n example translations by off-the-shelf search engine, and then we calculate the cosine similarity between their sentence embeddings and select the highest one as the matched example. Details will be introduced in the experiment section. Later, in order to easy to understand the similarity between the matched example and the source sentence, we also introduce the Fuzzy Match Score \cite{koehn2010convergence} as a measurement, which is computed as follows:
\begin{equation}
\setlength{\abovedisplayskip}{3pt}
\setlength{\belowdisplayskip}{3pt}
{\rm FMS}(x,x^m) =  1 - \frac{{\rm Levenshtein}(x, x^m)}{{\rm max}(|x|, |x^m|)}
\end{equation}

\begin{figure}[tt]
\centering
\includegraphics[scale=0.38]{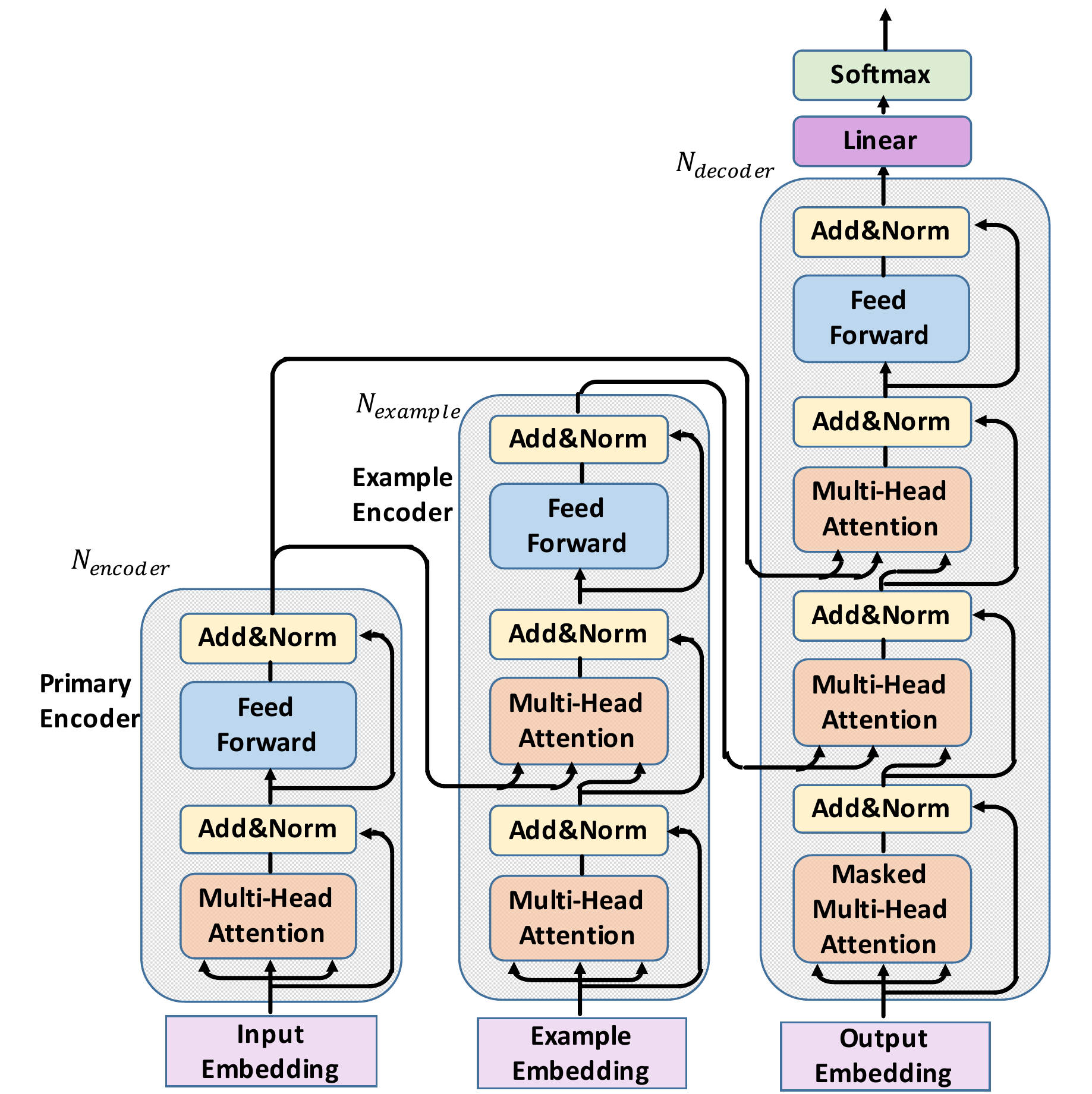}
\caption{Architecture of the basic model for EGNMT. Positional encodings are omitted to save space.}
\label{figure of Basic EGNMT Model Architecture}
\end{figure}

\subsection{Basic Model}

Figure \ref{figure of Basic EGNMT Model Architecture} shows the architecture for the basic model built upon the Transformer. We use two encoders: the primary encoder for encoding the source sentence $x$ and the example encoder for the matched example translation $y^m$. 
The primary encoder is constructed following Vaswani et al. \cite{vaswani2017attention}:
\begin{equation}
    \nu^{\rm src} = {\rm TransformerEncoder}(x)
\end{equation}
The example encoder contains three sub-layers: a multi-head example self-attention layer, a multi-head source-example
attention layer and a feed-forward network layer. Each sublayer is followed by a residual connection and layer normalization.

Before we describe these three sublayers in the example encoder, we first define the embedding layer. We denote the matched example translation as $y^m = [y_1^m, ..., y_L^m]$ where $L$ is the length of $y^m$. The embedding layer is then calculated as: 
\begin{equation}
    Y^{m} = [\hat{y}_1^{m}, ..., \hat{y}_L^{m}]
\end{equation}
\begin{equation}
    \hat{y}_j^{m} = {\rm Emb}(y_j^m) + {\rm PE}(j)
\end{equation}
where ${\rm Emb}(y_j^m)$ is the word embedding of $y_j^{m}$ and PE is the positional encoding function. 

The first sub-layer is a multi-head self-attention layer formulated as:
\begin{equation}
    A^{m} = {\rm MultiHead}(Y^{m}, Y^{m}, Y^{m})
\end{equation}
The second sub-layer is a multi-head source-example
attention which can be formulated as: 
\begin{equation}
    F^{m} = {\rm MultiHead}(A^{m}, \nu^{\rm src}, \nu^{\rm src})
\end{equation}
where $\nu^{src}$ is the output of the primary encoder. This sublayer is responsible for the attention between the matched example translation and the source sentence. The third sub-layer is a feed-forward network defined as follows:
\begin{equation}
    D^{m} = {\rm FFN}(F^{m})
\end{equation}



Different from the primary encoder with 6 layers, the example encoder has only one single layer. In our preliminary experiments, we find that a deep example encoder is not better than a single-layer shallow encoder. This may be due to the findings of recent studies, suggesting that higher-level representations in the encoder capture semantics while lower-level states model syntax \cite{peters2018deep,anastasopoulos2018tied,dou2018exploiting}. As the task is to borrow reusable fragments from the example translation, we do not need to fully understand the entire example translation. We conjecture that a full semantic representation of the example translation even disturbs the primary encoder to convey the meaning of the source sentence to the decoder.

\begin{figure}[tt]
\centering
\includegraphics[scale=0.36]{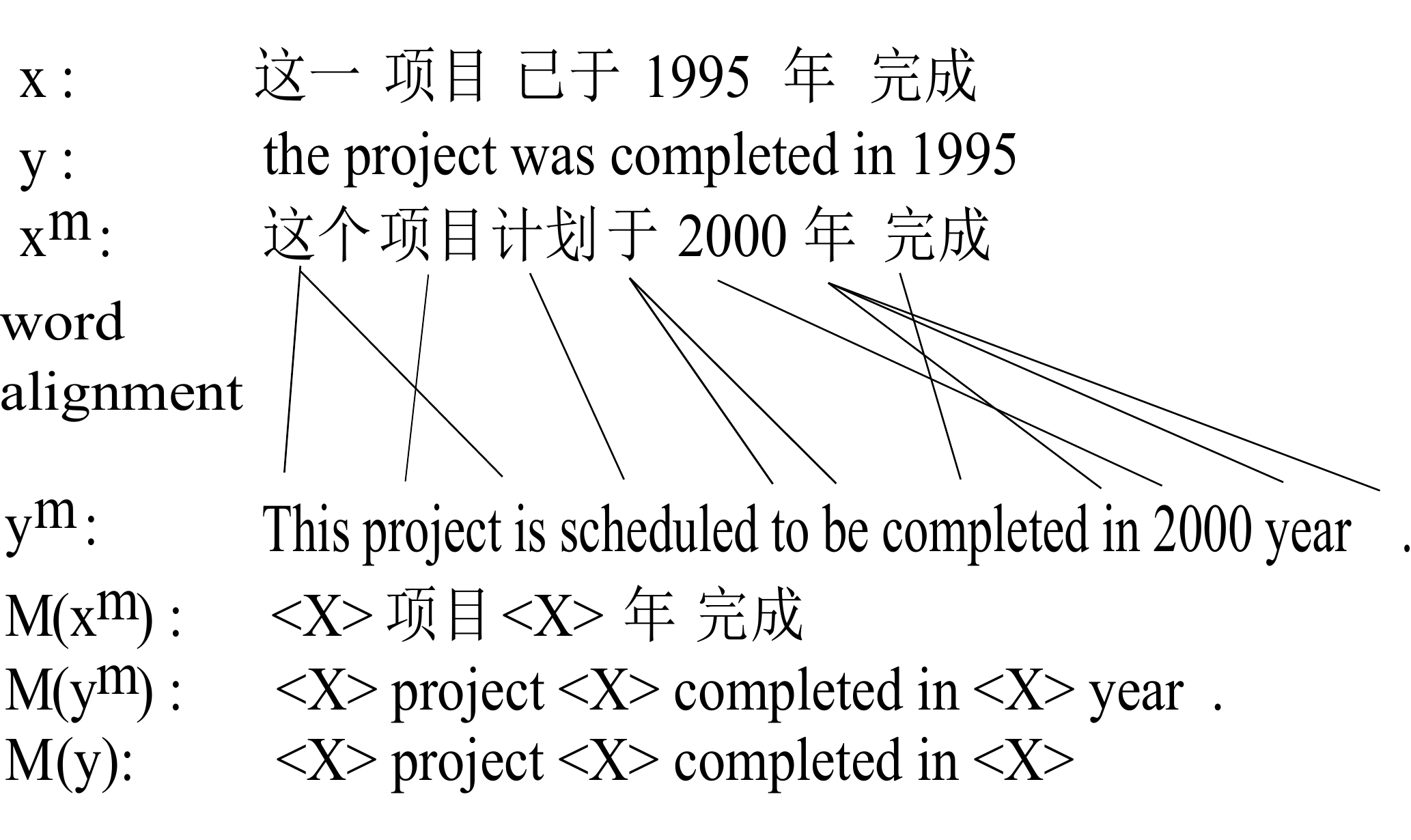}
\caption{An example demonstrating the masking process. }
\label{figure of Method Example}
\end{figure}

In the decoder, different from Vaswani et al. \cite{vaswani2017attention}, we insert an additional sub-layer between the masked multi-head self-attention and encoder-decoder attention. The additional sublayer is built for the attention of the decoder to the example translation representation:
\begin{equation}
\setlength{\abovedisplayskip}{3pt}
\setlength{\belowdisplayskip}{3pt}
H = {\rm Multihead}(\kappa, \nu^{\rm exp}, \nu^{\rm exp})
\end{equation}
where $\kappa$ is the output of the masked multi-head self-attention, and $\nu^{\rm exp}$ is the output of the example encoder. This sub-layer also contains residual connection and layer normalization.

\subsection{Noise-Masked Encoder Model}

\begin{figure}[!t]
\centering
\includegraphics[scale=0.36]{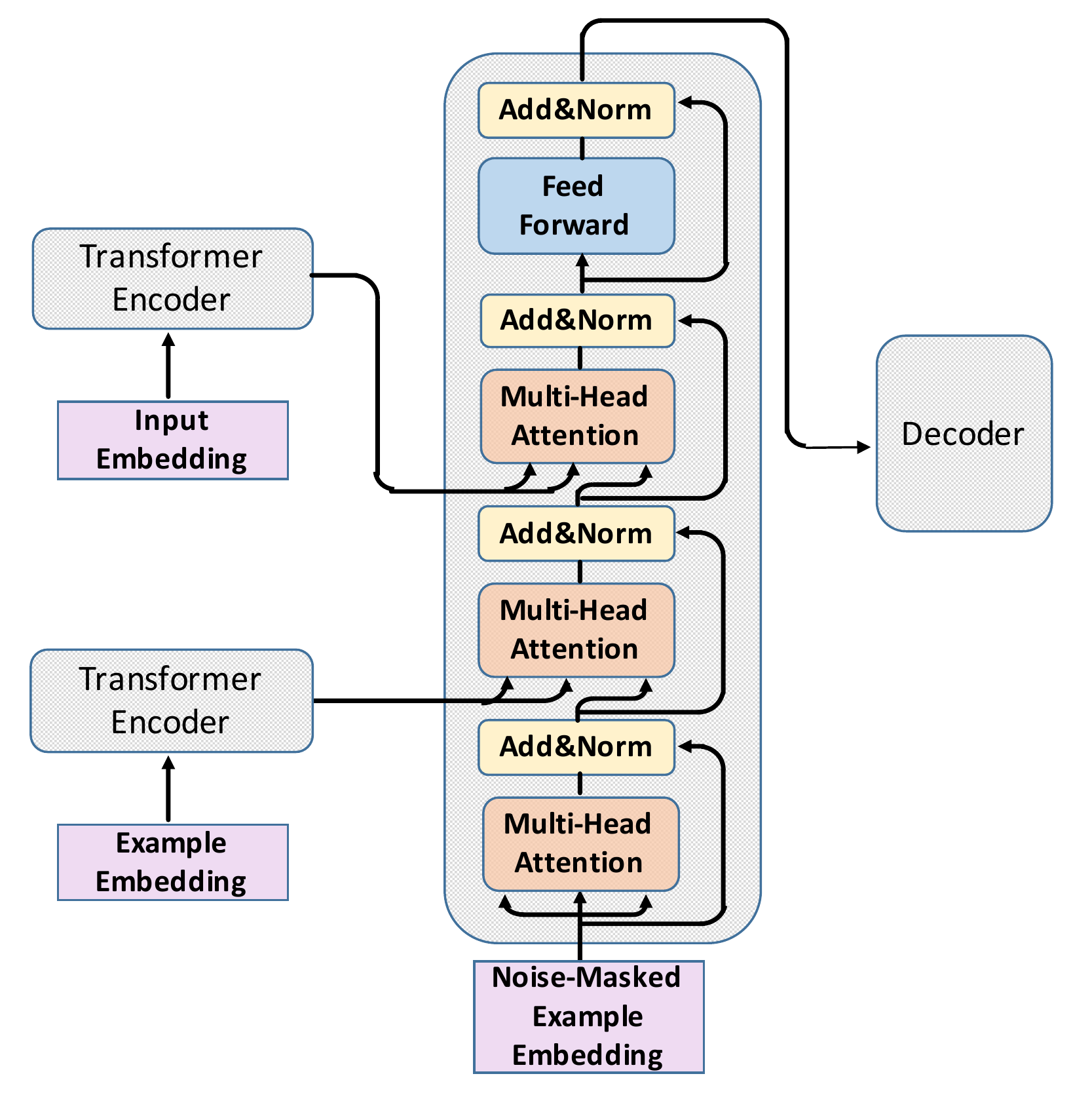}
\caption{Architecture of the NME model for EGNMT.}
\label{figure of NME model}
\end{figure}

As the source part of the matched example $x^m$ is not identical to the source sentence $x$, parts of the example translation cannot be reused in producing the target translation for $x$. These unmatched parts may act like noisy signals to disturb the translation process of the decoder. In order to prevent these unmatched parts from interrupting the target prediction, we propose a noise-masked encoder to encode the example translation. The idea behind this new encoder is simple. We detect the unmatched parts in the example translation and use a symbol ``$\langle X \rangle$'' to replace them so as to mask out their effect on translation. The masking process can be defined as a function $M$, from which we have the noise-masked example translation $M(y^m)$ from $y^m$. 

The masking function can be visualized with an example shown in Figure \ref{figure of Method Example}. Comparing the source side $x^m$ of the matched example with the source sentence, we can
find repeated source words. 
Keeping the repeated words and replacing other words with ``$\langle X \rangle$'', we obtain the masked version $M(x^m)$. Then, we use a pre-trained word alignment model to obtain word alignments between $x^m$ and $y^m$. We replace words in $y^m$
that are aligned to the masked parts in $M(x^m)$ with ``$\langle X \rangle$''. In this way, we finally obtain the masked example translation where only reusable parts are retained.

This masking method is based on word alignments. In practice, inaccurate word alignments will cause reusable words to be filtered out and noisy words retained. In order to minimize the negative impact of wrong word alignments as much as possible, we employ a standard transformer encoder module to encode the original example translation:
\begin{equation}
    \nu^{\rm oriexp} = {\rm TransformerEncoder}(y^m)
\end{equation}

Hence the differences between the example encoder in the basic model and NME model are twofold: (1) we replace the input $y^m$ with $M(y^m)$; (2) we add a sub-layer between the multi-head self-attention and source-example attention, to attend to the original example translation:
\begin{equation}
    K = {\rm MultiHead}(\iota, \nu^{\rm oriexp}, \nu^{\rm oriexp})
\end{equation}
where $\iota$ is the output of the multi-head self-attention.
The architecture can be seen in Figure \ref{figure of NME model}.

\subsection{Auxiliary Decoder Model}

In order to better leverage useful information in original example translations, we further propose an auxiliary decoder model. In this model, we directly compare the example translation $y^m$ with the target translation $y$. We can easily detect translation fragments that occur both in the example and real target translation. Similarly, we mask out other words to get a masked version $M(y)$ of the target translation (see the last row in Figure \ref{figure of Method Example}).

\begin{figure}[!t]
\centering
\includegraphics[scale=0.38]{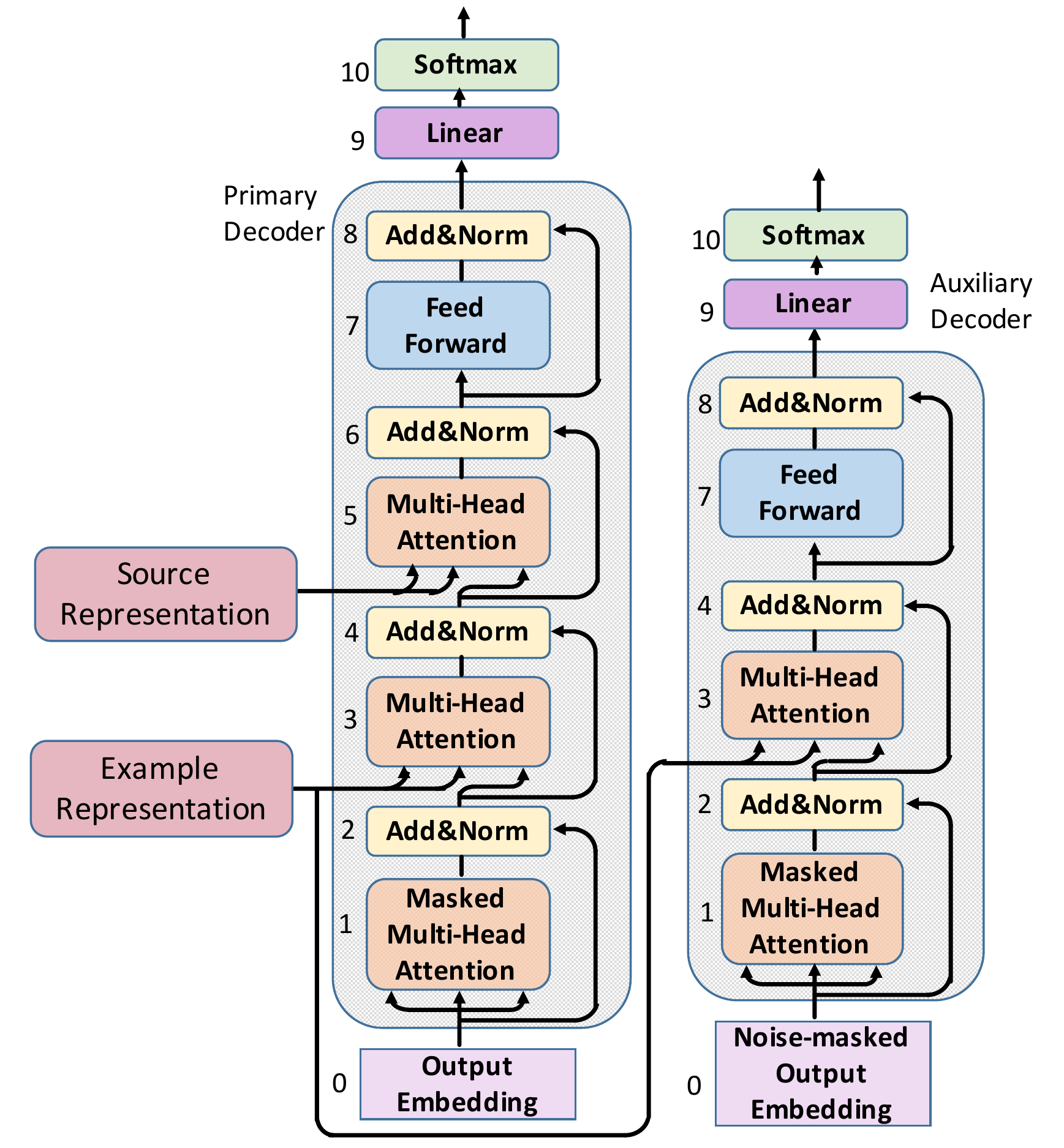}
\caption{Architecture of the auxiliary decoder model. The modules with the same indicator numbers in the primary and auxiliary decoder share parameters.}
\label{figure of AD Model}
\end{figure}

As the gold target translation $y$ is only available during the training phase, we employ an auxiliary decoder in the new model which is shown in Figure \ref{figure of AD Model}. The purpose for the auxiliary decoder is to predict the masked target translation $M(y)$ during the training phase from the example translation $y^m$ and $x$. It can be formulated as:
\begin{equation}
\setlength{\abovedisplayskip}{3pt}
\setlength{\belowdisplayskip}{3pt}
p(M(y)|x, y^m)= \prod p(M(y)_t|M(y)_{<t}, x, y^m)
\end{equation}

For this, we need to train an auxiliary NMT system with training instances $\{(x, y^m, M(y))\}$. The primary NMT system is trained with $\{(x, y^m, y)\}$. We jointly train these two systems to minimize a joint loss as follows:
\begin{equation}
\setlength{\abovedisplayskip}{3pt}
\setlength{\belowdisplayskip}{3pt}
L_{\rm joint} = L_{\rm pri} + L_{\rm aux}
\end{equation}
where $L_{\rm pri}$ is the loss for the primary NMT system while the latter $L_{\rm aux}$ is for the auxiliary NMT system.

During the testing phase, the auxiliary decoder is removed. We therefore share the parameters of the auxiliary decoder with the primary decoder. 
This is important as it allows the primary decoder to learn from the auxiliary decoder in the training phase to generate reusable parts. The joint training makes the primary decoder pay more attention to the reusable parts in the example translation by adjusting parameters in the attention network between the example encoder and the primary decoder to right directions.

\subsection{Assembling NME and AD}

The noise-masked encoder model and auxiliary decoder model can be combined together. In this assembling, we not only mask out noise parts in example translations in the encoder but also use the masked example translation to predict the masked target translation in the auxiliary decoder.

\section{Experiments}

We conducted experiments on Chinese-English, English-German and English-Spanish translation to evaluate the proposed models for EGNMT.

\begin{table*}[t]
\centering
\small
\caption{\label{BLEU scores in Chinese-English} BLEU scores for different models on Chinese-English translation. \#S: the number of sentences. T(all data): Transformer, TB: Basic EGNMT model, NME: Noise-Masked Encoder, AD: Auxiliary Decoder, Final: Basic model+Noise-Masked Encoder+Auxiliary Decoder, MET: matched example translations, Gated: method from Cao and Xiong \cite{cao2018encoding}}
\begin{tabular}{c|c|c|c|c|c|c|c|c}
\toprule
\bf FMS & \bf \#S & \bf T(all data) & \bf TB & \bf TB+NME & \bf TB+AD & \bf Final & \bf MET & \bf Gated \\ \hline
$[0.9,1.0)$ & 171 & 55.71 & 69.72 & 66.47 & 86.58 & 88.17 & \textbf{94.23} & 79.95  \\
$[0.8,0.9)$ & 182 & 63.09 & 72.61 & 69.36 & 83.49 & \textbf{86.96} & 79.84 & 81.45  \\
$[0.7,0.8)$ & 178 & 62.56 & 69.67 & 67.63 & 78.55 & \textbf{79.40} & 67.11 & 74.37  \\
$[0.6,0.7)$ & 179 & 67.76 & 71.17 & 71.08 & 77.04 & \textbf{77.31} & 58.93 & 71.22  \\
$[0.5,0.6)$ & 181 & 69.21 & 70.63 & 70.06 & 72.75 & \textbf{73.53} & 46.99 & 69.91 \\
$[0.4,0.5)$ & 177 & 74.28 & 72.97 & 74.13 & 74.10 & \textbf{74.50} & 34.67 & 73.14 \\
$[0.3,0.4)$ & 180 & 68.39 & 66.46 & \textbf{68.58} & 66.97 & 66.85 & 22.93 & 65.92 \\
$[0.2,0.3)$ & 185 & 50.57 & 48.67 & \textbf{52.58} & 49.96 & 50.97 & 9.72 & 49.79 \\
$(0.0,0.2)$ & 181 & \textbf{35.43} & 32.05 & 35.35 & 33.53 & 34.66 & 1.18 & 32.16  \\ 
\hline
\hline
$(0.0,1.0)$ & 1,614 & 60.07 & 62.85 & 63.12 & 68.53 & \textbf{69.94} & 47.32 & 66.93 \\ 
\hline
\hline
\#Param  & - & 92M & 102M & 102M & 102M & 102M & - & 104M \\
\bottomrule
\end{tabular}
\end{table*}

\subsection{Experimental Settings}

We implemented our example-guided NMT systems based on Tensorflow. We obtained word alignments with the tool fast-align\footnote{Available at: https://github.com/clab/fast\_align}. The maximum length of training sentences is set to 50 for all languages. We applied byte pair encoding \cite{sennrich2016neural} with 30k merging operations. We used the stochastic gradient descent algorithm with Adam \cite{kingma2015adam} to train all models. We set the beam size to 4 during decoding. We used two GPUs for training and one for decoding. We used case-insensitive 4-gram BLEU as our evaluation metric \cite{papineni2002bleu} and the script ``multi-bleu.perl'' to compute BLEU scores.

For Chinese-English corpus, we used the United National Parallel Corpus \cite{rafalovitch2009united} from Cao and Xiong \cite{cao2018encoding}, which consists of official records and other parliamentary documents. The numbers of sentences in the training/development/test sets are 1.1M/804/1,614. 

We also experimented our methods on English-German and English-Spanish translation. We used the JRC-Acquis corpus\footnote{Available at https://ec.europa.eu/jrc/en/language-technologies/jrc-acquis} following previous works \cite{koehn2010convergence,gu2017search,bapna2019non}. We randomly selected sentences from the corpus to build the training/development/test sets. The numbers of sentences in the training/development/test sets for English-German are 0.5M/676/1,824 and 0.8M/900/2,795 for English-Spanish. We used the training sets as the example database.
We firstly used the Lucene\footnote{Available at http://lucene.apache.org/} to retrieve top 10 example translations from the example database excluding the sentence itself. Then we obtained the sentence embeddings of these retrieved examples with the fasttext tool\footnote{Available at: https://fasttext.cc/} and calculated the cosine similarity between the source sentence and each retrieved example. Finally we selected the example with the highest similarity score as the matched example.

\subsection{Chinese-English Results}

Table \ref{BLEU scores in Chinese-English} shows the results. In the table, we divide the test set into 9 groups according to the FMS values of matched example translations and show BLEU scores on each group and the entire set. We show the BLEU scores for both the baseline and matched example translations against reference translations for comparison.
Additionally, we adapted the gated method proposed by Cao and Xiong \cite{cao2018encoding} to the Transformer and compared with this gated Transformer model. The results of this experiment are also reported in Table \ref{BLEU scores in Chinese-English}. From the table, we can observe that
\begin{itemize}

\item The basic model obtains an improvement of 2.78 BLEU points over the baseline. This demonstrates the advantage of example-guided NMT: teaching NMT to learn from example translations on the fly is better than mixing examples as training data. We also find that the basic model can improve translation quality only when FMS is larger than 0.5, indicating that it suffers from noises in low-FMS example translations.

\item The noise-masked encoder model is better than the basic model by 0.27 BLEU points. The model significantly improves translation quality for sentences with low-FMS example translations, which means that masking noise is really helpful. But it also slightly hurts translation quality for high-FMS (e.g., $>$0.5) sentences compared with the basic model. This may be because the noisy parts are much more dominant than the reusable parts in example translations with low FMS, which makes easier to detect and mask out noisy parts via word alignments. However, in high-FMS example translations, many words can be reused with a few unmatched words scattered in them. It is therefore risky to detect and mask out reusable words with inaccurate word alignments. Although we also attend to the original example translation, reusable words that are masked mistakenly may still not be replenished.

\item The auxiliary decoder model hugely improves the performance by more than 5.68 BLEU points over the basic model. It significantly improves translation quality for high-FMS sentences by learning to reuse previously translated segments separated by scattered unmatched words. However, in low FMS intervals, its performance is still not satisfactory for that they may not distinguish the unmatched parts accurately.

\item Assembling the noise-masked encoder and auxiliary decoder models together, we achieve the best performance, 7.09 BLEU points higher than the basic model and 3.01 BLEU points than the previous gated Transformer model \cite{cao2018encoding}. We can improve translation quality for both high-FMS and low-FMS sentences. This is because, on the one hand, we can mask the noisy information in the example by the NME model, on the other hand, through the AD model, we can learn to let the model use the useful information. The AD model can also guide the NME model in the attendance to the original example.

\end{itemize}

\subsection{Results for English-German and English-Spanish Translation}

We further conducted experiments on the English-German and English-Spanish corpus. Results are shown in Table \ref{BLEU scores for en-de} and \ref{BLEU scores for en-es}. We have similar findings to those on Chinese-English translation. Our best model achieves improvements of over 4 BLEU points over the basic EGNMT model. The improvements in these two language pairs are not as large as those in Chinese-English translation. The reason may be that the retrieved examples are not as similar to German/Spanish translations as those to English translations in the Chinese-English corpus. This can be verified by the BLEU scores of matched example translations in Chinese-English, English-German and English-Spanish corpus, which are 47.32/36.51/38.72 respectively. The more matched example translations are similar to target translations, the higher improvements our model can achieve.

\begin{table}
\centering
\small
\caption{\label{BLEU scores for en-de} BLEU scores of EGNMT on English-German translation.}
\begin{tabular}{c|c|c|c|c|c}
\toprule
\bf FMS & \bf \#S & \bf T(all data) & \bf TB & \bf Final & \bf MET \\ \hline
$[0.9,1.0)$ & 199 & 68.07 & 77.91 & 82.26 & \textbf{83.69} \\
$[0.8,0.9)$ & 210 & 63.02 & 70.43 & \textbf{72.89} & 68.33 \\
$[0.7,0.8)$ & 205 & 62.20 & 66.23 & \textbf{69.62} & 61.43 \\
$[0.6,0.7)$ & 203 & 58.02 & 59.17 & \textbf{63.88} & 51.89 \\
$[0.5,0.6)$ & 207 & 57.65 & \textbf{62.44} & 62.30 & 44.55 \\
$[0.4,0.5)$ & 183 & 52.34 & 51.97 & \textbf{56.49} & 32.83 \\
$[0.3,0.4)$ & 205 & 48.72 & 45.36 & \textbf{51.43} & 23.47 \\
$[0.2,0.3)$ & 206 & 43.15 & 40.1 & \textbf{44.49} & 16.47 \\
$(0.0,0.2)$ & 206 & \textbf{37.49} & 32.02 & 37.03 & 6.35 \\ 
\hline
\hline
$(0.0,1.0)$ & 1,824 & 54.19 & 55.85 & \textbf{59.25} & 36.51  \\
\bottomrule
\end{tabular}
\end{table}

\begin{table}
\centering
\small
\caption{\label{BLEU scores for en-es} BLEU scores of EGNMT on English-Spanish translation.}
\begin{tabular}{c|c|c|c|c|c}
\toprule
\bf FMS & \bf \#S & \bf T(all data) & \bf TB & \bf Final & \bf MET \\ \hline
$[0.9,1.0)$ & 367 & 66.94 & 68.38 & 79.94 & \textbf{80.43} \\
$[0.8,0.9)$ & 363 & 68.63 & 69.57 & \textbf{73.20} & 66.30 \\
$[0.7,0.8)$ & 364 & 68.18 & 69.38 & \textbf{71.42} & 54.12 \\
$[0.6,0.7)$ & 363 & 68.68 & 69.45 & \textbf{70.21} & 48.11 \\
$[0.5,0.6)$ & 274 & 62.04 & \textbf{62.91} & 62.79 & 32.94 \\
$[0.4,0.5)$ & 161 & 58.41 & \textbf{58.71} & 58.02 & 28.96 \\
$[0.3,0.4)$ & 230 & 58.29 & 57.06 & \textbf{61.91} & 24.09\\
$[0.2,0.3)$ & 343 & 53.48 & 53.98 & \textbf{54.02} & 15.36 \\
$(0.0,0.2)$ & 330 & 49.68 & 49.80 & \textbf{50.47} & 9.53 \\ 
\hline
\hline
$(0.0,1.0)$ & 2,795 & 60.31 & 60.90 & \textbf{64.35} & 38.72  \\
\bottomrule
\end{tabular}
\end{table}

\begin{table}
\centering
\small
\caption{\label{right and wrong words in example} The numbers of matched and unmatched noisy words in example translations. O: original matched example translations. M: noise-masked example translations. n: noisy words. m: matched words. }
\begin{tabular}{c|c|c|c|c}
\toprule
\bf FMS & \bf O(m) & \bf O(n) & \bf M(m) & \bf M(n)  \\ \hline
$[0.0,0.2)$ & 148 & 2,404 & 99 & 265 \\
$[0.2,0.3)$ & 476 & 1,739 & 380 & 190 \\
$[0.3,0.4)$ & 1,007 & 1,370 & 893 & 225 \\
$[0.4,0.5)$ & 1,251 & 1,227 & 1,146 & 205 \\
$[0.5,0.6)$ & 1,559 & 885 & 1,410 & 228 \\
$[0.6,0.7)$ & 2,029 & 740 & 1,888 & 210 \\
$[0.7,0.8)$ & 2,154 & 536 & 1,987 & 155 \\
$[0.8,0.9)$ & 2,340 & 352 & 2,210 & 116 \\
$[0.9,1.0)$ & 2,424 & 100 & 2,294 & 33 \\ 
\hline
\hline
$(0.0,1.0)$ & 13,388 & 9,353 & 12,307 & 1,627 \\
\bottomrule
\end{tabular}
\end{table}

\begin{table*}[t] 
\centering
\small
\caption{\label{Example} A translation sample from the test set. Reusable parts are highlighted in bold.} 
\begin{tabular}{l||l}
\toprule
source & feizhoudalu de wuzhuangchongtu , genyuan daduo yu pinkun ji qianfada youguan . \\
\hline
reference & \textbf{most armed conflicts} on the african continent \textbf{are rooted in poverty} and under-development . \\
\hline 
$x^m$ & youguan feizhou guojia de wuzhuangchongtu , jiu@@ qi@@ genyuan daduo yu pinkun he qianfada youguan . \\
\hline
$y^m$ & \textbf{most armed conflicts} in and among african countries \textbf{are rooted in poverty} and lack of development . \\
\hline
Transformer & \textbf{most of the armed conflicts} on the continent \textbf{are related to poverty} and the less developed countries . \\
\hline
Basic Model & \textbf{most armed conflicts} in the african continent \textbf{are related to poverty} and lack of development . \\
\hline
Final model & \textbf{most armed conflicts} in the african continent \textbf{are rooted in poverty} and lack of development . \\
\bottomrule
\end{tabular}
\end{table*}

\section{Analysis}

We look into translations generated by the proposed EGNMT models to analyze how example translations improve translation quality in this section.

\subsection{Analysis on the Generation of Reusable Words}

We first compared matched example translations against reference translations in the Chinese-English test set at the word level after all stop words are removed. Table \ref{right and wrong words in example} shows the number of matched and unmatched noisy words in example translations. The noise-masking procedure can significantly reduce the number of noisy words (9,353 vs. 1,627). 8.1\% of matched words in the original example translations are filtered out due to wrong word alignments. 

We collected a set of reusable words $R$ that are present in both example and reference translations (all stop words removed). Similarly, we obtained a set of words $S$ that occur in both example and system translations. The words in $S$ can be regarded as words generated by EGNMT models under the (positive or negative) guidance of example translations. The intersection of $R$ and $S$ is the set of words that are correctly reused from example translations by EGNMT models.  We computed an F$_1$ metric for reusable word generation as follows:
\begin{equation}
\setlength{\abovedisplayskip}{3pt}
\setlength{\belowdisplayskip}{3pt}
\begin{split}
    p = |R\cap S| / |S|  \qquad 
    r = |R\cap S| / |R| \\
    F_1 = 2*p*r / (p+r) \qquad \quad
\end{split}
\end{equation}

\begin{figure}[!t]
\centering
\includegraphics[scale=0.35]{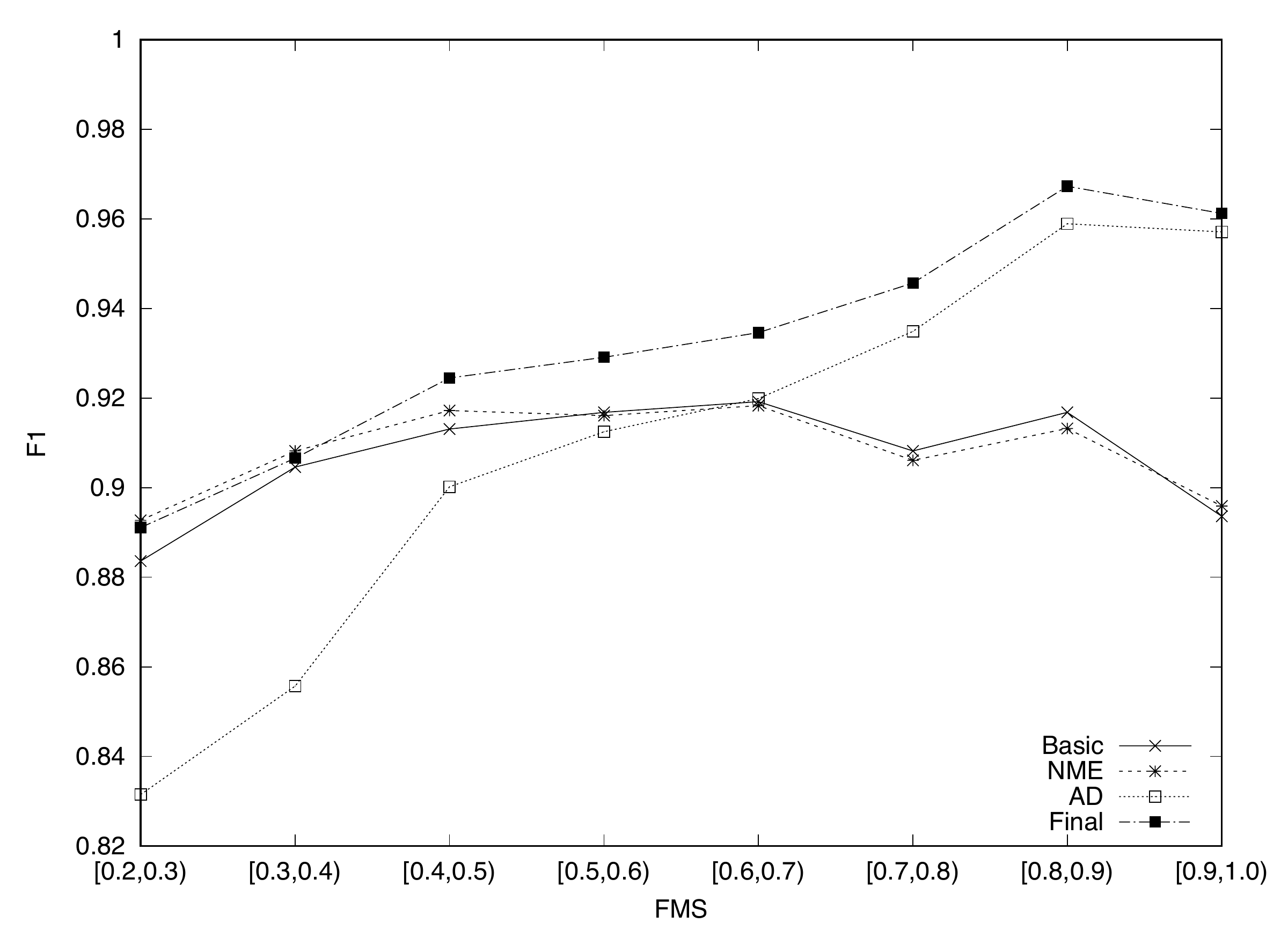}
\caption{Reusable word generation F1 scores of EGNMT models.}
\label{figure of F1 scores}
\end{figure}

Figure \ref{figure of F1 scores} shows the F$_1$ scores for different EGNMT models. It can be seen that the proposed EGNMT models is capable of enabling the decoder to generate matched words from example translations while filtering noisy words.

The reason that the auxiliary decoder model achieves the lowest F$_1$ for low-FMS sentences is because the model reuses a lot of noisy words from low-FMS example translations (hence the precision is low). This indicates that low-FMS example translations have a negative impact on the AD model. The NME model is able to achieve a high precision by masking out noisy words but with a low recall for high-FMS examples by incorrectly filtering out reusable words. Combining the strengths of the two models, we can achieve high F$_1$ scores for both low- and high-FMS examples as shown in Figure \ref{figure of F1 scores} (the final model).

\subsection{Attention Visualization and Analysis}

\begin{figure}[tt]
\centering
\includegraphics[scale=0.35]{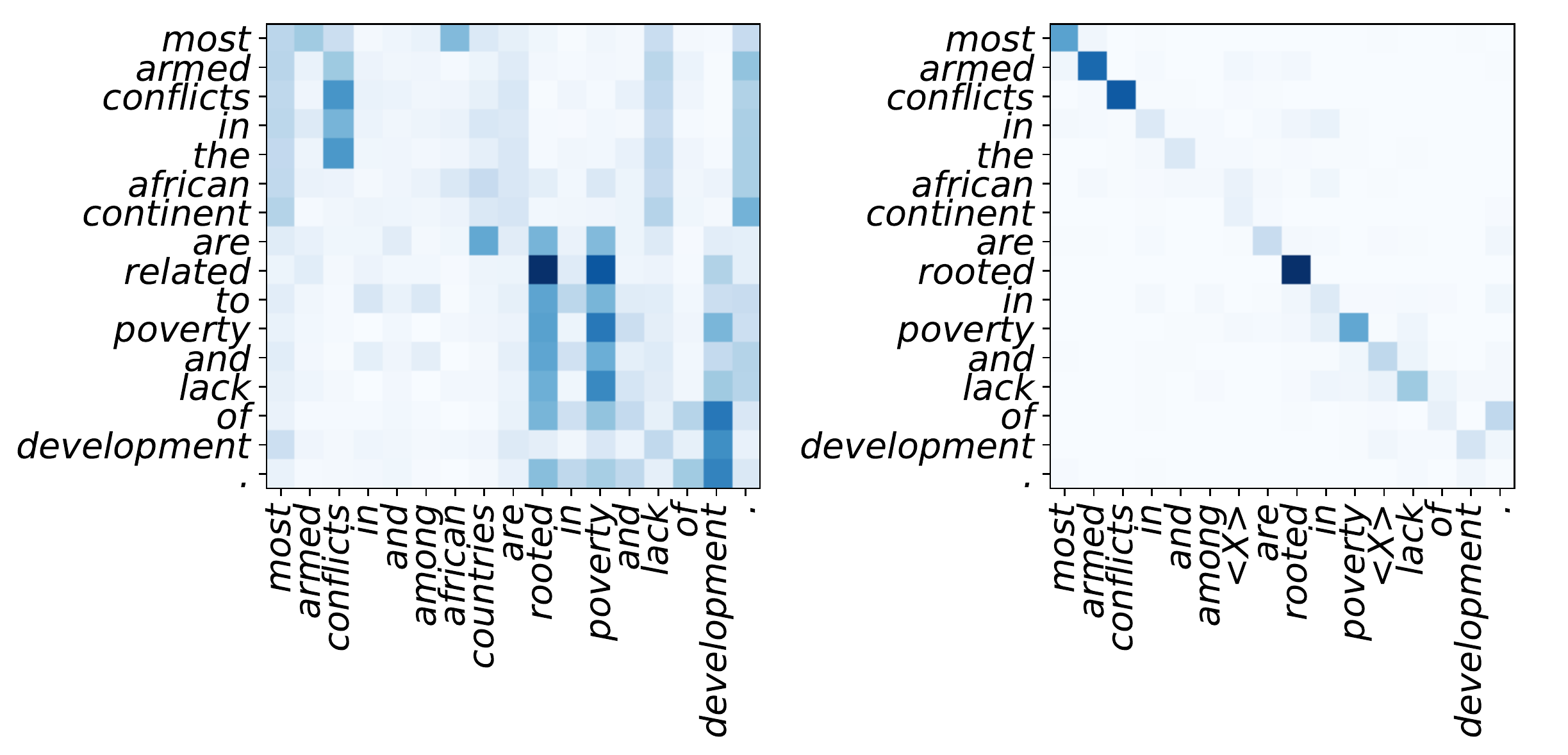}
\caption{Visualization of attention weights between the example translation (X-axis) and the system translations (Y-axis) generated by the basic model (left) and final model (right).}
\label{figure about attention}
\end{figure}

Table \ref{Example} provides a sample from the Chinese-English test set.
We can see that the example translation provides two fragments that are better than the target translation generated by the baseline model. The fragment ``most armed conflicts" is successfully reused by the basic model, but the fragment ``are rooted in poverty" does not appear in the target translation generated by the basic model. In contrast to the two models, our final model successfully reuses the two fragments.

We further visualize and analyze attention weights between the example translation and system translation (the example encoder vs. the primary decoder). The visualization of attention weights for this sample is shown in Figure \ref{figure about attention}. Obviously, the basic EGNMT model can use only a few reusable words as the attention weights scatter over the entire example translation rather than reusable words. The final EGNMT system that uses both the noise-masked encoder and auxiliary decoder model, by contrast, correctly detects all reusable words and enables the decoder to pay more attention to these reusable words than other words.

\section{Conclusions and Future Work}

In this paper, we have presented EGNMT, a general and effective framework that enables the decoder to detect and take reusable translation fragments in generated target translations from the matched example translations. The noise-masking technique is introduced to filter out noisy words in example translations. The noise-masking encoder and auxiliary decoder model are proposed to learn reusable translations from low- and high-FMS example translations. Both experiments and analyses demonstrate the effectiveness of EGNMT and its advantage over mixing example translations with training data.


It is natural to use the proposed EGNMT to combine NMT with translation memory. 
Although not explored, the EGNMT framework can also be used to adapt an NMT system to a domain with very little in-domain data by treating the in-domain data as the example database. 
In addition to these, there are still many open problems related to the use and abstraction of examples in neural machine translation that have not been addressed in this paper. In particular, our work can be extended in the following two ways.

\begin{itemize}

\item Example-based NMT with multiple examples. We only use the best example with the highest similarity score.
However, we can find many examples with reusable fragments.
A seamless combination of example-based translation philosophy with NMT is necessary for example-based NMT to benefit
from multiple examples.
\item Integration of translation rules or templates into NMT. The noise-masked example in this paper can be considered as a template abstracted from a single example. If we learn translation templates or rules from multiple examples or the entire training data, the noise-masked encoder model or the auxiliary decoder model might be adjusted to incorporate them into NMT.
\end{itemize}

\bibliography{ecai}
\end{document}